# L-ReLF: A Framework for Lexical Dataset Creation

Anass Sedrati
*Department of Computer Science*
*KTH, Royal Institute of Technology*
Stockholm, Sweden
anass@kth.se

Mounir Afifi*
*Wikimedia Morocco*
Rabat, Morocco
prebirthtime@gmail.com
*Corresponding author

Reda Benkhadra
*Wikimedia Morocco*
Rabat, Morocco
m.benkhadra@aui.ma

*Abstract*—This paper introduces the L-ReLF (Low-Resource Lexical Framework), a novel, reproducible methodology for creating high-quality, structured lexical datasets for underserved languages. The lack of standardized terminology, exemplified by Moroccan Darija, poses a critical barrier to knowledge equity in platforms like Wikipedia, often forcing editors to rely on inconsistent, ad-hoc methods to create new words in their language. Our research details the technical pipeline developed to overcome these challenges. We systematically address the difficulties of working with low-resource data, including source identification, utilizing Optical Character Recognition (OCR) despite its bias towards Modern Standard Arabic, and rigorous post-processing to correct errors and standardize the data model. The resulting structured dataset is fully compatible with Wikidata Lexemes, serving as a vital technical resource. The L-ReLF methodology is designed for generalizability, offering other language communities a clear path to build foundational lexical data for downstream NLP applications, such as Machine Translation and morphological analysis.

## I. Introduction

The digital age has amplified the academic and technological dominance of high-resource languages, notably English, creating a significant linguistic disparity that hinders global knowledge equity [1, 2]. This challenge is particularly acute for underrepresented, unstandardized languages, which often lack the formal, structured lexical resources necessary to articulate modern scientific and technical concepts [3].

In Morocco, Darija refers to various used forms of dialectal Arabic that share common features. While spoken by an overwhelming majority in the country (either as native tongue or second language), it remains seen only as an oral way of communication, while Standard Arabic or French are preferred officially for writing and taught in schools.

Within the Wikimedia ecosystem, this deficiency directly impedes the growth and quality of projects like Wikipedia. Editors working in languages such as Moroccan Darija—our paper's case study—are forced to rely on ad-hoc methods for terminology creation, leading to widespread inconsistencies, frustrating edit conflicts, and ultimately, hampering the development of comprehensive and trustworthy knowledge bases. An additional complexity comes from the fact that most editors are volunteer non-subject matter experts, who do not have the necessary tools in terms of linguistics to conduct such complex tasks. On the other hand, established academics do not participate in big numbers in Wikimedia projects (mostly for the claim of its lack of credibility) [4], making their important perspectives and knowledge missing in the biggest encyclopedia in the world [5].

To address this critical gap, we introduce L-ReLF (Low-Resource Lexical Framework), a novel, reproducible, and accessible methodology for generating high-quality lexical datasets for underserved languages. Our primary goal is to empower linguistic volunteer communities with the foundational tools needed for language standardization and digital growth, without needing to be experts in linguistics. By combining rigorous linguistic requirements with computational data engineering, the L-ReLF is designed to move terminology creation beyond subjective, isolated efforts toward a sustainable, community-driven technical pipeline.

This paper's specific contribution is the detailed technical description of the L-ReLF methodology for data collection and preparation. We present the systematic, step-by-step pipeline used to convert disparate physical sources into a cohesive, structured, and Wikidata-compatible dataset for Moroccan Darija. This framework is crucial because the quality and structure of the initial data determine the success of all downstream NLP applications. We detail the technical challenges encountered, such as working with imperfect Optical Character Recognition (OCR) and integrating web-scraped data and present our solutions for data standardization and linguistic structuring. This work culminates in the creation of a robust dataset, serving as a vital technical resource.

The remainder of this paper is structured as follows: Section II reviews related work concerning the challenges of Low-Resource Language Processing (LRLP) and the utility of Wikidata Lexemes as a data infrastructure. Section III presents the core of our contribution: the L-ReLF Methodology, detailing Source Identification, the Data Gathering and Extraction Pipeline, and the rigorous Data Preparation and Standardization process. Finally, Section IV discusses the technical utility of the resulting dataset for future computational work (e.g., Machine Translation and Morphological Analysis) and emphasizes that the full methodology, including the developed scripts and data model, is publicly available and reproducible by other low-resource language (LRL) communities.

This research work was funded through The Wikimedia Research Fund, supported by the Wikimedia Foundation.

## II. Related Work and Background

### A. The Shift from Rule-Based to Corpus-Based LRLP

Research in Low-Resource Language Processing (LRLP) has historically struggled with a "data divide," where languages like English benefit from vast, annotated resources, while languages like Moroccan Darija suffer from acute data sparsity, making it difficult to train robust computational models for essential tasks like machine translation, morphological analysis, or named entity recognition [6]. This data deficit is compounded by several linguistic and infrastructural issues:

- **Lack of Standardization:** Many underserved languages, including Moroccan Darija, lack standardized orthography and formal linguistic bodies to regulate term creation, resulting in significant dialectal variation and spelling inconsistency within available text [7]. This lack of standardization directly impacts the creation of usable datasets, as data are typically noisy and require extensive manual cleaning. This effort is often overlooked in general research, yet it is the most critical barrier to building robust resources

- **Limited Resources:** Computational tools like part-of-speech taggers, parsers, and comprehensive electronic dictionaries are often non-existent or of poor quality. When resources do exist, they are frequently trapped in non-digital formats, such as print dictionaries.

- **Global Linguistic Barrier:** This technical resource gap is sustained by the global barrier posed by English dominance in academia and technology, which diverts funding and research effort away from LRLPs. This disparity hinders scientific progress and severely limits knowledge equity for speakers of LRLs [1,2,3].

A synthesis of recent literature reveals a shift in addressing this gap. Early attempts, such as [8], relied on rule-based approaches to generate vocabulary. However, these methods presuppose the existence of high-quality root lexicons, which are rarely available in digital formats for dialects.

Conversely, more recent initiatives have pivoted toward creating massive, unstructured text corpora in Moroccan Darija. Works by [9, 10] have successfully aggregated large datasets for machine translation and chat-based Large Language Models (LLMs), while [11] have bridged gaps in sentiment analysis. While these contributions are critical for statistical modeling, they lack the fine-grained grammatical tagging required for lexical standardization. They prioritize token quantity over structured quality (morphology, root derivation, gender), leaving a gap in the resources needed for dictionary generation and deep morphological analysis

L-ReLF differentiates itself by targeting lexical data: a structured, standardized collection of information about the words (lexemes) themselves, including their grammatical properties, morphological derivations, senses, and semantic relationships [12]. Unlike a corpus, this structure enables the extraction of morphological patterns required for systematic new term generation.

For our project, aiming to create a methodology for new term generation, this structured level of detail is paramount because:

- Linguistic Pattern Extraction: It provides the necessary, granular data to analyze existing patterns of terminology generation, and systematically create new, grammatically correct words.

- Computational Utility: It forms the essential input for high-level NLP applications like morphological analyzers and structured dictionary generation, tasks a raw text corpus cannot directly support.

### B. The Infrastructure Gap: From Text to Knowledge Graphs

There is a distinct difference between a text corpus and a structured lexical knowledge base. Established high-resource infrastructures, such as [13], provide a blueprint for organizing lexical concepts, yet these infrastructures remain largely empty for underrepresented languages. [14] demonstrates the potential of Wikidata Lexemes as an open, structured solution for languages like Danish. Wikidata Lexemes offer a highly structured, multilingual environment that can store fine-grained grammatical and semantic data, making them ideal for developing new terminology in a scalable and community-auditable manner. However, literature lacks a synthesized framework for migrating a low-resource language into such infrastructures when the source material is non-digital.

L-ReLF addresses this specific intersection by shifting focus from "collecting text" to "structuring lexemes" compatible with Linked Open Data standards. Currently, Moroccan Darija volunteer Wikipedia editors use ad-hoc methods—relying on manual translation or personal judgment— that often fail within this structured environment, leading to the inconsistencies and edit conflicts that our systematic methodology seeks to overcome.

### C. Data Collection Techniques for LRLs

Building lexical resources for languages where information is primarily stored in print requires specialized and carefully adapted data collection techniques. The current state-of-the-art relies primarily on two core techniques: Optical Character Recognition (OCR) for print resources and web-scraping for existing digital materials.

- **OCR Limitations:** While powerful for high-resource languages, the accuracy of commercial OCR tools significantly degrades when applied to LRLs that use complex scripts, such as Arabic. Commercial OCR tools are predominantly trained on Modern Standard Arabic (MSA) and frequently hallucinate corrections, altering valid Darija syntax to fit MSA norms, both in terms of letters and morphology. [8]. This means that data gathered via OCR requires substantial, costly manual correction and validation before it can be used for any computational task, a limitation we address explicitly in our methodology.

- **Web-scraping and data standardization:** While web-scraping helps acquire digital texts, the resulting data is typically unstructured and inconsistent. To be useful for computational tasks, it must be subjected to extensive cleanup and standardization to fit the required format.

While current literature acknowledges these problems, it does not propose a standardized workflow to mitigate it. Participatory research in African languages, such as that by the Masakhane community, emphasizes the necessity of "Human-in-the-Loop" (HITL) methodologies for dataset creation [15]. Our L-ReLF framework synthesizes these findings by formalizing a pipeline where OCR serves only as a preliminary step, mandating a rigorous human verification process to ensure the linguistic integrity required for the technical downstream tasks required later on.

III. METHODOLOGY: DATASET CREATION AND STRUCTURING

The creation of a high-quality lexical dataset for a LRL requires a technical methodology that prioritizes rigor, standardization, and replicability over mere quantity. This section details the L-ReLF approach, which transforms fragmented, non-digital linguistic data for Moroccan Darija into a structured, machine-readable format suitable for computational processing.

*A. Source Identification and Selection*

The initial phase focused on identifying and selecting sources that minimize data noise and maximize linguistic richness.

Our primary criteria for source selection were:

1. **Script and Format:** Sources must be in Arabic script (the standard for formal technical use, distinguishing it from Latin-script social media data).

2. **Specialization and Relevance:** Sources had to contain specialized and technical vocabulary rather than general, colloquial words, ensuring fitness for the project's goal of terminology creation.

3. **Authority and Availability:** Sources had to be academic dictionaries or published linguistic texts available in physical or digital formats.

We focused exclusively on print resources, ultimately gathering data from eight academic Darija books[1], with a target scope of approximately **5,000** distinct words. We intentionally avoided using web-scraping for digital sources (social media, forums) due to severe time constraints, the high degree of unstandardized Latin script usage, and the fact that such unstructured text corpora do not contain the necessary lexical data (morphology, gender, etymology) that is central to our framework. For LRL methodologies, we recommend prioritizing structured book sources, even non-digitally, as they offer cleaner data requiring less complex preprocessing than raw web text.

*B. Data Gathering and Extraction Pipeline*

The data gathering process had to contend with the non-digital nature of the selected sources, relying heavily on transcription and digitalization.

The main technique employed was Optical Character Recognition (OCR) to convert scanned documents into text format. We opted for Google Drive's built-in OCR due to its accessibility, simplicity, and efficiency in handling bulk document conversion, which offers a favorable time-saving and good quality trade-off [16], despite its known limitations with LRLs [17]. While more complex, specialized OCR tools exist, our decision prioritized a reproducible, low-barrier solution suitable for adoption by other resource-constrained language communities.

The extraction pipeline revealed immediately three main technical and linguistic limitations inherent to LRL data digitalization:

1. **Poor Scan Quality and Manual Input:** Some sources, notably the dictionaries by Amīlī, had an inherently poor print quality in their original physical versions. This poor-quality rendered OCR unusable, necessitating full manual transcription for approximately 20% of the corpus.

2. **Diglossic OCR Bias:** The OCR tool, primarily trained on MSA, systemically misinterpreted Darija-specific letters and often attempted to correct Darija words to match MSA orthography. This introduced significant errors and false assumptions into the raw output, necessitating extensive post-processing. This issue is unfortunately present in all OCR solutions in Arabic script, as none of them supports specifically Darija.

3. **Methodological Diversity of Sources:** The selected books, while authoritative, used different linguistic methodologies and internal structures. Some sources provided long descriptive text for word explanations, while others offered only simple translations or tables. Key lexical fields, such as etymology and grammatical gender, were an integral part of some sources, but completely absent in others. This required a unified consolidation approach and a consequent work of manual input, a critical challenge when comprehensive source material and time are scarce.

---

[1] The eight selected books were the following (presented in original name in Arabic, followed by Latin alphabet transcription using the British Standard, and English translation):
1. *Muḥammed šafīq: Addāriǧa almaġribiyya, maǧālu tawārud bayna al amazīġīa wa al ʿarabiyya* (1999)
   Mohammed Chafik: Moroccan Darija, area of convergence between Berber and Arabic
2. *Muḥammed Ben Al Bašīr Būsellām: Muʿǧam addāriǧa almaġribiyya: Al ǧuḏūr wa al iḫtilāfāt al ǧihawiyya - Al ǧuzʾu Al Awwal (2015) - Attānī (2018) - Attālit (2024)*
   Mohammed Ben Al Bachir Bousellam: Dictionary of Moroccan Darija: Roots and Regional Differences - Part One (2015) - Two (2018) - Three (2024)
3. American Peace Corps: Moroccan Arabic Textbook (2011)
4. *Markaz tanmiyyat addārīǧa zagūra: Qāmūs addāriǧa almaġribiyya (2017)*
   Zakoura Darija Development Center: Dictionary of Moroccan Darija
5. *Ḥasan Amīlī: Al muʿǧam al baḥrī wa al milāḥī (2011), muʿǧam al bināa wa al miʿmār (2014)*
   Hassam Amili: Maritime and Navigation Dictionary (2011), Construction and Architecture Dictionary (2014)

## C. Data Preparation and Standardization

This phase, implementing a dual-pass verification protocol, is the most resource-intensive, focusing on transforming the raw, inconsistent input into a single, standardized, and structured lexical dataset, meeting the rigorous Wikidata standards.

First, a thorough manual and semi-automated cleanup process is essential to overcome the OCR-induced errors and structural inconsistencies. This involved a visual sanity check where the generated text was manually verified against the original source book to correct character-level errors (e.g., substituting 'ۀ' with 'ك'). This time-consuming task was necessary because the numerous errors and inherent assumptions made by the OCR engine would otherwise have compromised the linguistic integrity of the dataset.

Then, to unify the disparate source data, a standardized spreadsheet model was applied. The key data fields were chosen for both linguistic relevance and suitability for Wikidata Lexemes, ensuring that the data would be machine-readable and ready for the final integration step. The fields include: *Darija word, Lexical category, Grammatical gender* and *Etymology*. During this phase, duplicates are removed. In fact, since data come from different and unrelated sources, it is not uncommon to find the same word mentioned more than once, sometimes in different spellings, making the manual verification even more crucial.

Another challenge in this step was that missing data in some sources (such as the grammatical gender field) required significant manual input from the authors based on native linguistic expertise.

Finally, explicit relationships between words were established within the dataset. This step went beyond simple data entry by capturing semantic and morphological connections (e.g., derivational links between verbs and nouns). This internal structuring is vital as it highlights the scientific rigor of the dataset, providing the explicit patterns necessary for the methodology creation phase to follow. Moreover, derivations provide a simple and straightforward way to enrich the dataset, with interrelated words that already have several fields in common.

## D. Quantitative Error Analysis and Quality Assurance

To address the lack of transparency regarding OCR performance and final dataset reliability, we conducted a quantitative sampling of the pipeline's accuracy.

1. **Quantification of OCR Error Rates:** A quantitative analysis of a random sample of ($N=100$) entries processed via Google Drive OCR revealed a Character Error Rate (CER) of 17%. Crucially, 59% of these errors were attributed to MSA bias, where the engine incorrectly 'corrected' Darija-specific graphemes to match Standard Arabic morphology. This high error rate necessitated a shift from automated ingestion to a semi-automated pipeline, where manual verification accounted for approximately 90% of the total dataset preparation time. This quantitative finding validates our assertion in Section II that fully automated extraction is currently unfeasible for dialectal Arabic and necessitates the Human-in-the-Loop (HITL) approach described in this framework.

2. **Final Dataset Accuracy:** The accuracy of the final dataset is defined as the fidelity of transcription relative to the authoritative physical sources. Given the dual-pass verification protocol, the final dataset achieves a transcription accuracy of 100% relative to the source material for the verified entries. By shifting the burden of accuracy from the algorithmic layer (OCR) to the verification layer (Human Experts), L-ReLF ensures that the "noise" typically associated with LRL datasets is eliminated before publication. The final dataset is not a probabilistic output of a model, but a deterministic digitization of authoritative linguistic scholarship.

## E. Wikidata Integration and Interoperability

The prepared dataset is particularly designed to be integrated into Wikidata Lexemes. This platform is chosen because it offers a structured, accessible, multilingual, and interoperable format that aligns with the principles of open research and language preservation. An advantage of Wikidata is that established tools such as QuickStatements and OpenRefine can be used to manage bulk data upload. It also allows the L-ReLF framework to be easily reproducible and sustainable as the software scripts developed to map the standardized spreadsheet columns to the required Wikidata Lexeme properties can be used again for all languages.

The L-ReLF workflow (figure 1 below) summarizes our methodology and steps for dataset creation. To handle the variance in source quality quantified in Section III, the framework splits inputs into two streams: an OCR-based path for high-quality scans and a manual transcription path for degraded physical sources. These streams converge at the Data Preparation step, acting as the unification layer. This step represents the Human-in-the-Loop (HITL) aspect, performing the dual-pass verification described in Section III-C. First, a visual sanity check corrects OCR diglossic bias, before a manual mapping of grammatical gender and etymology is performed if this data is missing in the original sources. Lastly, the final output phase shows the mapping of the standardized dataset into the Wikidata Lexeme Schema. In this step the flat tabular data is converted into a linked open data graph, establishing the specific semantic relationships required for the downstream NLP tasks.

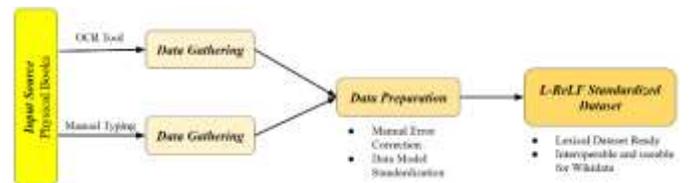

Fig. 1. L-ReLF Workflow: From Printed Source to Structured Lexical Dataset

## IV. Conclusion, Utility and Future Work

This paper presented L-ReLF (Low-Resource Lexical Framework), by detailing a lexeme-centric methodology for creating high-quality datasets in underserved languages. By outlining a systematic pipeline for data gathering via adapted OCR and manual input, and rigorous standardization, we established a technical framework that directly addresses the deficiencies inherent in the previous ad-hoc methods used for languages like Moroccan Darija.

The structured dataset and its associated scripts hold substantial technical utility for future computational work. This resource provides the necessary, linguistically-annotated foundation for advancing Natural Language Processing (NLP) research in a resource-scarce environment. Specifically, the data is engineered to enable the training and fine-tuning of machine translation (MT) models and the development of computational tools for morphological analysis. Furthermore, the standardized format, optimized for the Wikimedia ecosystem, facilitates the automation of digital dictionaries (such as the Wiktionary) and the expansion of Wikidata Lexemes.

Despite the methodological rigor, several limitations were encountered during the data creation phase. The primary technical constraint was the OCR bias and MSA confusion, which required substantial time to resolve due to the tool's tendency to incorrectly "adjust" Darija words to fit Modern Standard Arabic conventions. This issue, combined with the extreme challenge of source structural inconsistency across the various dictionaries, necessitated lengthy manual intervention. Furthermore, the reliance on manual typing for low-quality physical sources and the need for error correction by manual verification proved to be time-consuming steps in the entire process, limiting the size of the final dataset. These challenges highlight the need for investment in LRL-specific OCR tools and robust, semi-automated data cleaning scripts in future iterations of this framework to reduce dependence on costly human labor.

This methodology is inherently reproducible and generalizable to other low-resource languages because its core steps are language-agnostic yet designed to adapt to LRL constraints. The framework explicitly accounts for the technical difficulty of sourcing materials by providing a clear protocol for combining OCR and manual data entry, a common necessity when working with non-digitized LRL sources. The data model standardization—the defining feature of L-ReLF—is based on universal linguistic categories (e.g., lexical category, etymology, semantic relationships) and the structured format of Wikidata. Any community can apply these identical steps and use the associated scripts, without being expert linguists, merely by substituting their LRL's specific data sources, ensuring the framework's broad adaptability.

The future phases of this work will aim to move from resource creation to application. This includes pattern extraction from the structured data, methodology creation for new term generation, and validation through a community survey. By clearly defining these stages, we keep the focus of the current paper on the technical contribution of dataset creation. The final data model and the technical scripts developed for integration are being made publicly available, guaranteeing that this technical contribution can be adapted by other global communities of volunteers.

ACKNOWLEDGMENT

This research was made possible through the financial support of the Wikimedia Foundation. The authors are particularly grateful to the Wikimedia Research Fund for awarding Grant ID: G-RS-2504-18952 which provided essential support for data collection, linguistic analysis, and community outreach. The views expressed in this paper are those of the authors and do not necessarily reflect the official policy or position of the Wikimedia Foundation.